\documentclass[journal,12pt,draftcls,onecolumn]{IEEEtran}
\usepackage[latin9]{inputenc}
\usepackage{url}
\usepackage{amsmath}
\usepackage{amssymb}
\usepackage{graphicx}
\providecommand{\U}[1]{\protect \rule{.1in}{.1in}}
\begin{document}

\title{Sky Highway Design for Dense Traffic}
\author{Quan~Quan, Mengxin Li\thanks{Q. Quan and M. Li are with the School of Automation Science and Electrical
Engineering, Beihang University, Beijing 100191, China (e-mail: qq\_buaa@buaa.edu.cn (Q. Quan)).}}
\maketitle

\begin{abstract}
The number of Unmanned Aerial Vehicles (UAVs) continues to explode. Within the
total spectrum of Unmanned Aircraft System (UAS) operations, Urban Air
Mobility (UAM) is also on the way. Dense air traffic is getting ever closer to
us. Current research either focuses on traffic network design and route design
for safety purpose or swarm control in open airspace to contain large volume
of UAVs. In order to achieve a tradeoff between safety and volumes of UAVs, a
sky highway with its basic operation for Vertical Take-Off and Landing (VTOL)
UAV is proposed, where traffic network, route and swarm control design are all
considered. In the sky highway, each UAV will have its route, and an airway
like a highway road can allow many UAVs to perform free flight. The
geometrical structure of the proposed sky highway with corresponding flight
modes to support dense traffic is studied one by one. The effectiveness of the
proposed sky highway is shown by the given demonstration.

\end{abstract}

\section{INTRODUCTION}

With the on-going miniaturization of motors, sensors and processors, the
number of Unmanned Aerial Vehicles (UAVs) continues to explode. UAVs
increasingly play an integral role in many applications in scenarios from the
on-demand package delivery to traffic and wildlife surveillance, inspection of
infrastructure, search and rescue, agriculture, and cinematography
\cite{Balakrishnan(2018)}. Within the total spectrum of Unmanned Aircraft
System (UAS) operations in the National Airspace System (NAS), Urban Air
Mobility (UAM) referring to flight operations to carry personnel and cargo
within the geographical limits of an urban metropolis are also on the way
\cite{NASA},\cite{SESAR}.

Traditionally, the main role of Air Traffic Management (ATM) is to keep
prescribed separation between all aircraft by using centralized control.
However, it is infeasible for the increasing number of UAVs as the traditional
control method is in lack of scalability. In order to address such problem,
free flight is a developing air traffic control method that uses decentralized
control \cite{FreeFlight(1995)},\cite{Jana(1997)}. By using Automatic
Dependent Surveillance-Broadcast (ADS-B) \cite{ICAO(2012)}, Vehicle to Vehicle
(V2V) communication or 5G mobile network \cite{Hayat(2016)}, UAVs can receive
the information of their neighboring UAVs to avoid collision. Especially in
low-altitude airspace, utilizing the existing mobile networks will eliminate
the need to deploy a new infrastructure and therefore, help to ensure feasible
connected UAVs economically \cite{GSMA(2018)}.

Air traffic for UAVs starts to draw the attention of more and more researchers
\cite{IoD(2016)},\cite{Devasia(2016)},\cite{Mcfadyen(2017)}. A concept, coined
as the Internet of Drones (IoD), was proposed in \cite{IoD(2016)}, where a
conceptual model of how its architecture can be organized was proposed and the
features of an IoD system were described. In \cite{Devasia(2016)}, based on a
proposed pre-established route network, an open-loop UAV operation paradigm
was proposed to enable a large number of relatively low-cost UAVs to fly
beyond the line of sight. In \cite{Mcfadyen(2017)}, a concept of network
design for a UAS Traffic Management (UTM) system was proposed, where feasible
low-level urban airspace regions, candidate network nodes and Unmanned Traffic
Network (UTN) structure were proposed one by one. Inspired by a swarm in
nature, dense traffic study for UAVs is also getting the momentum. In
\cite{Viragh(2016)}, some different dense multirotor UAV traffic simulation
scenarios in open 2D and 3D space were studied under realistic environments in
the presence of sensor noise, communication delay, limited communication
range, limited sensor update rate, and finite inertia. Furthermore, in
\cite{Boldizsar(2018)}, a general, decentralized air traffic control solution
using autonomous drones was proposed, where some difficult dense traffic
situations were challenged based on the force-based distributed multi-robot
control model. On the one hand, some air traffic studies focus on traffic
network design and route design, following the idea of the traditional air
traffic transport but with unique differences and challenges. Based on
designed traffic networks, the risk to people on the ground will be reduced
greatly. On the other hand, some air traffic studies stemmed from swarm
control \cite{Chung(2018)} often focus on the free flight in open airspace. By
using swarm control, larger volumes of UAVs can fly at ease in the same airspace.

Motivated by these current studies, a trade-off scheme, called \emph{sky
highway}, is proposed for VTOL UAVs, in which traffic network, route and swarm
control design are all considered. The sky highway is a traffic network on
which any VTOL UAV should have its own route. Based on this, risk control can
be eased. Unlike aforementioned networks, UAVs in an airway(a straight-line
segment of a route between two nodes) are controlled in a swarm control
manner, where the airway like a highway road can allow many UAVs to realize
free flight. In this way, the volume of UAVs is increased.

In this paper, the geometrical structure of airways and intersections in the
network with corresponding flight modes are designed and studied one by one to
support dense traffic. Concretely,

\begin{itemize}
\item The geometrical structure of the network rather than abstract lines and
nodes is designed to maintain a certain distance between two UAVs on two
carriageways (Contribution i).

\item A highway mode is proposed for the airway flight, where a finishing line
rather than several waypoints are proposed to avoid potential traffic jam
(Contribution ii).

\item Also, the highway mode is applied to a type of intersections. For
another type of intersection, a rotary island mode is proposed. With this two
modes, the UAV traffic flow can be increased greatly (Contribution iii).
\end{itemize}

\section{PROBLEM FORMULATION}

In this section, a traffic network model is introduced first. Then, objective
of this paper is proposed.

\subsection{General Traffic Network Model}

\begin{figure}[b]
\centering
\includegraphics[scale= 1]{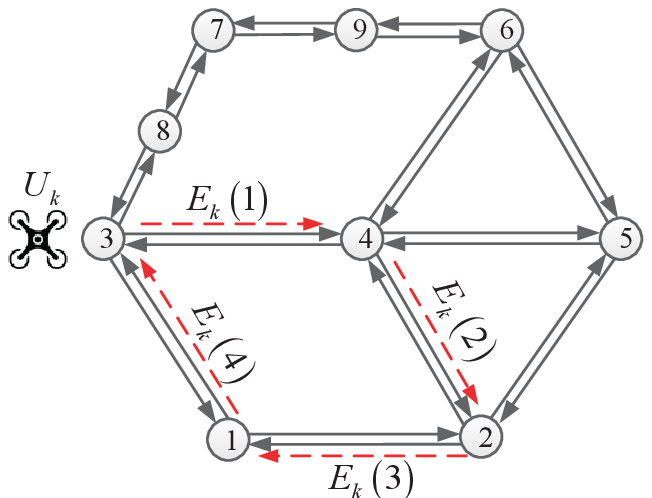} \caption{Traffic network}%
\label{network}%
\end{figure}

The airspace is structured similar to the roads network in the cities. Because
of the characteristic of the airspace, the airlines could be as shown in Fig.
\ref{network}. A general traffic network model can be formulated by using the
graph theory \cite{Devasia(2016)}. Similar to the idea of \cite{IoD(2016)},
UAVs are only allowed inside the following three kinds of areas:
\emph{airways} which play a similar role to the roads, \emph{intersection
nodes} which connect at least two airways, \emph{termination nodes} which
connect at least one airway where UAVs fly in and out of the network.

\subsection{Objective}

Some basic principles are provided to guide the traffic network design.

\begin{itemize}
\item (i) Every VTOL UAV has its route\emph{.}

\item (ii) UAVs should not collide with each other, so they should be
separated at a certain distance. With such a distance $r_{\text{a}}>0$, two
UAVs do not need to take extra effort to avoid each other.

\item (iii) UAVs should not fly outside of airways.

\item (iv) Every UAV can have different velocities with a setting turning radius.

\item (v) UAVs can be allocated different priorities.
\end{itemize}

Based on these basic principles, an air traffic network, named \emph{sky
highway}, is designed for \emph{dense traffic}. The dense traffic is denser
than the current civil aviation transportation. It is similar to the traffic
in highways for cars, which allows many UAVs to fly in the same airway and
then to pass an intersection. Obviously, the traffic network general model
shown in Fig. \ref{network} is too abstract for the dense traffic. So, based
on this general model, the objective of this paper is to further design the
structure of airways and intersections and include as well their corresponding
flight modes which can support dense traffic. For simplicity, only all airways
and intersections on the same altitude are focused on.

\section{AIRWAY AND INTERSECTION DESIGN FOR DENSE TRAFFIC}

In this section, the structure of airways and intersections with corresponding
flight modes for supporting dense traffic are studied one by one.

\subsection{Airway Design and Flight Mode}

\subsubsection{Airway Design}

\begin{figure}[b]
\centering
\includegraphics[scale= 1]{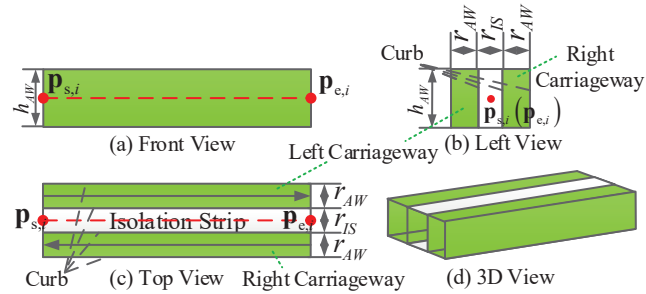} \caption{Airway structure}%
\label{airwaystrucure}%
\end{figure}

An airway is designed as shown in Fig. \ref{airwaystrucure}. The $i$th
\emph{airway} denoted by\emph{ }$W_{\left[  \mathbf{p}_{\text{s},i}%
,\mathbf{p}_{\text{e},i}\right]  }$ is a set of cuboid (length $l_{\left[
\mathbf{p}_{\text{s},i},\mathbf{p}_{\text{e},i}\right]  }>0,$ width
$2r_{AW}+r_{IS}>0,$ height $h_{AW}>0$), which is divided into three parts, two
\emph{carriageways} for traffic traveling in opposite directions, namely
$W_{\left[  \mathbf{p}_{\text{s},i},\mathbf{p}_{\text{e},i}\right]
,\text{left}}$, $W_{\left[  \mathbf{p}_{\text{s},j},\mathbf{p}_{\text{e}%
,j}\right]  ,\text{right}}$ (length $l_{\left[  \mathbf{p}_{\text{s}%
,i},\mathbf{p}_{\text{e},i}\right]  }>0,$ width $r_{AW,i}>0,$ height
$h_{AW}>0$), separated by an isolation strip also in the form of a cuboid,
namely, $W_{\left[  \mathbf{p}_{\text{s},j},\mathbf{p}_{\text{e},j}\right]
,\text{iso}}$ (length $l_{\left[  \mathbf{p}_{\text{s},i},\mathbf{p}%
_{\text{e},i}\right]  }>0,$ width $r_{IS}>0,$ height $h_{AW}>0$). The line
segment $\left[  \mathbf{p}_{\text{s},i},\mathbf{p}_{\text{e},i}\right]  $ is
the center line on the middle height of the airway, where $\mathbf{p}%
_{\text{s},i},\mathbf{p}_{\text{e},i}\in%
%TCIMACRO{\U{211d} }%
%BeginExpansion
\mathbb{R}
%EndExpansion
^{3}$ are the starting point and the ending point, respectively. So,
\[
l_{\left[  \mathbf{p}_{\text{s},i},\mathbf{p}_{\text{e},i}\right]
}=\left \Vert \mathbf{p}_{\text{s},i}-\mathbf{p}_{\text{e},i}\right \Vert .
\]
Airways should be designed such that\emph{ }two UAVs in two different
carriageways are separated at a distance $r_{\text{a}}>0.$ This implies two
UAVs on two carriageways do not need to avoid with each other. Here, two
different carriageways may be in one airway or two different airways. Before
proceeding, the distance between two sets is defined. The distance between set
$S_{1}\ $and set$\ S_{2}$ is defined as%
\[
d\left(  S_{1},S_{2}\right)  \triangleq \underset{\mathbf{x}\in S_{1}%
,\mathbf{y}\in S_{2}}{\min}\left \Vert \mathbf{x-y}\right \Vert .
\]

(a) When two different carriageways are in the same airway, the isolation
strip plays an important role. A result is given in \textit{Proposition 1}.

\textbf{Proposition 1}. Suppose the airway structure is shown in Fig.
\ref{airwaystrucure}. For the $i$th airway, if
\begin{equation}
r_{IS}>r_{\text{a}} \label{p1con}%
\end{equation}
then%
\begin{equation}
d\left(  W_{\left[  \mathbf{p}_{\text{s},i},\mathbf{p}_{\text{e},i}\right]
,\text{left}},W_{\left[  \mathbf{p}_{\text{s},i},\mathbf{p}_{\text{e}%
,i}\right]  ,\text{right}}\right)  >r_{\text{a}}. \label{p1result}%
\end{equation}

(b) When two different carriageways are in two different airways, the two
airways should be separated at a distance $r_{a}>0.$ A result is given in
\textit{Proposition 2}\textbf{.}

\textbf{Proposition 2}. Suppose the airway structure is shown in Fig.
\ref{airwaystrucure}. For the $i$th and the $j$th airways, if
\begin{equation}
d\left(  \left[  \mathbf{p}_{\text{s},i},\mathbf{p}_{\text{e},i}\right]
,\left[  \mathbf{p}_{\text{s},j},\mathbf{p}_{\text{e},j}\right]  \right)
>r_{\text{a}}+\sqrt{h_{AW}^{2}+\left(  2r_{AW}+r_{IS}\right)  ^{2}}
\label{p2con}%
\end{equation}
then%
\begin{equation}
d\left(  W_{\left[  \mathbf{p}_{\text{s},i},\mathbf{p}_{\text{e},i}\right]
},W_{\left[  \mathbf{p}_{\text{s},j},\mathbf{p}_{\text{e},j}\right]  }\right)
>r_{\text{a}},i\neq j. \label{p2result}%
\end{equation}

\begin{figure}[b]
\centering
\includegraphics[scale= 1]{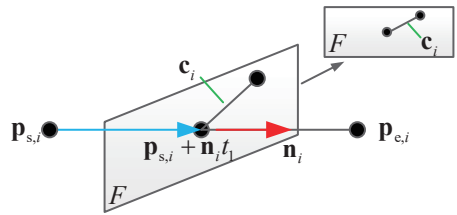} \caption{Position representation
in an airway}%
\label{linedist}%
\end{figure}\textit{Proof}. As shown in Fig. \ref{linedist}, let
$\mathbf{p}_{i}\in W_{\left[  \mathbf{p}_{\text{s},i},\mathbf{p}_{\text{e}%
,i}\right]  }\ $and $\mathbf{p}_{j}\in W_{\left[  \mathbf{p}_{\text{s}%
,j},\mathbf{p}_{\text{e},j}\right]  }$ be written as
\begin{align}
\mathbf{p}_{i}\left(  t_{1},\mathbf{c}_{i}\right)   &  =\mathbf{p}%
_{\text{s},i}+\mathbf{n}_{i}t_{1}+\mathbf{c}_{i}\label{add1}\\
\mathbf{p}_{j}\left(  t_{2},\mathbf{c}_{j}\right)   &  =\mathbf{p}%
_{\text{s},j}+\mathbf{n}_{j}t_{2}+\mathbf{c}_{j} \label{add2}%
\end{align}
Here, $\mathbf{n}_{i}=\frac{\mathbf{p}_{\text{e},i}-\mathbf{p}_{\text{s},i}%
}{\left \Vert \mathbf{p}_{\text{e},i}-\mathbf{p}_{\text{s},i}\right \Vert },$
$\mathbf{n}_{j}=\frac{\mathbf{p}_{\text{e},j}-\mathbf{p}_{\text{s},j}%
}{\left \Vert \mathbf{p}_{\text{e},j}-\mathbf{p}_{\text{s},j}\right \Vert }$,
$t_{1}\in \left[  0,\left \Vert \mathbf{p}_{\text{e},i}-\mathbf{p}_{\text{s}%
,i}\right \Vert \right]  $, $t_{2}\in \left[  0,\left \Vert \mathbf{p}%
_{\text{e},j}-\mathbf{p}_{\text{s},j}\right \Vert \right]  $ and $\mathbf{c}%
_{i},\mathbf{c}_{j}\in F\subset%
%TCIMACRO{\U{211d} }%
%BeginExpansion
\mathbb{R}
%EndExpansion
^{2}$, where $F$ is a cross section of the airway perpendicular to line
segment $\left[  \mathbf{p}_{\text{s},i},\mathbf{p}_{\text{e},i}\right]  $
shown in Fig. \ref{linedist}.

Based on the relation (\ref{add1}) and (\ref{add2}), the conclusion can be
drawn. $\square$

\subsubsection{Flight Mode}

\begin{figure}[t]
\centering
\includegraphics[scale= 1]{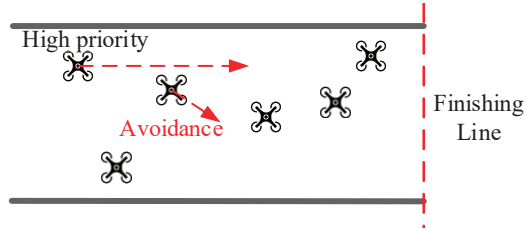} \caption{Highway mode}%
\label{AirwayDesign}%
\end{figure}

For dense traffic, a \emph{highway mode} is proposed. A carriageway has
several lanes but does not require UAVs on lanes one by one like trains on a
railway. UAVs under decentralized control are attracted by a finishing line at
the end of the airway rather than several waypoints (as shown in Fig.
\ref{AirwayDesign}). Once approaching the finishing line, it will switch to
another airway, for example, via an intersection. So, the principle (i) is
satisfied. UAVs should repel each other if their distance is too short, for
example, by using the artificial potential field method. Also, the curbs (as
shown in Fig. \ref{airwaystrucure}) are required to repel UAVs to keep them
within the carriageway. So principles (ii) and (iii) are satisfied. If UAVs
have different velocities, then the fast UAV and the slow UAV will interact
with each other according to a certain protocol. For example, a protocol is
based on the artificial potential field method. Then the fast UAV and slow UAV
will often change their flight directions rather than only slow down the fast
one. Generally, the fast one will overtake the slow one. Then, principle (iv)
is satisfied. If a UAV with a high priority has to overtake the slow UAV, then
the UAV with a high priority can keep the same velocity, but the UAVs with
lower priorities should make avoidance. 

The highway mode is suitable for dense traffic for the following reasons. (i)
The airspace is utilized sufficiently. (ii) Existent finishing line will avoid
the traffic jam at the end of the airway because the velocities of UAVs are
aligned like birds flying in the sky \cite{Dutta(2010)}. (iii) In order to
perform overtaking, the fast one can often frequently change its direction a
little bit to avoid the slow one rather than only reduce its speed.

\subsection{Intersection Design and Flight Mode}

\subsubsection{Intersection Classification}

\begin{figure}[b]
\centering
\includegraphics[scale= 1]{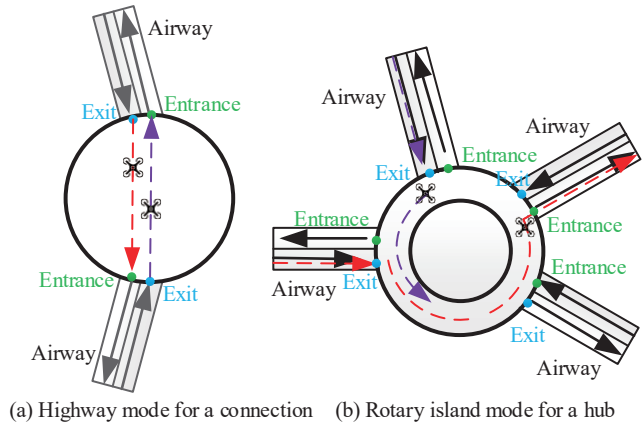} \caption{Modes for two
types of intersections}%
\label{intersectionnew}%
\end{figure}

An \emph{intersection} connects at least two airways. As shown in Fig.
\ref{intersectionnew}, there are two cases.

\begin{itemize}
\item (i) An intersection, called a \emph{connection }or\emph{ azimuth}
\emph{connection}, is connected with two airways to change azimuth, because
only all airways and intersections on the same altitude are focused on. For
example, Nodes $1,7,8,9$\emph{ }are shown in Fig. \ref{network}.

\item (ii) An intersection is used as a hub through which UAVs can select
different airways to go. It connects with more than two airways, called\emph{
}a\emph{ hub}, shown in Fig. \ref{intersectionnew}(b),\emph{ }such as Nodes
$2,4,5,6$ shown in Fig. \ref{network}. It is required that all airways connect
with a hub intersection.
\end{itemize}

In the following, some basic principles on intersections are described.
Furthermore, two modes for the two types of intersections are proposed.

\subsubsection{Intersection Design}

\begin{figure}[t]
\centering
\includegraphics[scale= 1]{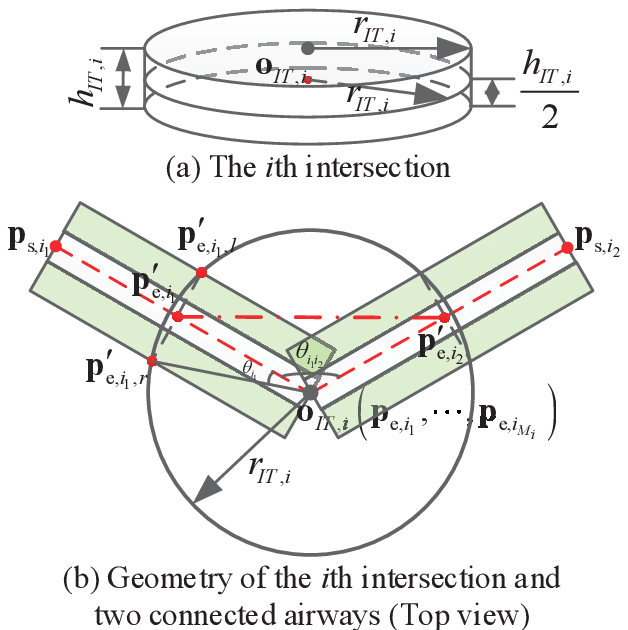} \caption{The
$i_{1}$th and $i_{2}$th airways are connecting with a hub intersection}%
\label{Intersection}%
\end{figure}

For the $i$th intersection, as shown in Fig. \ref{Intersection}, suppose the
center lines $\left[  \mathbf{p}_{\text{s},i_{j}},\mathbf{p}_{\text{e},i_{j}%
}\right]  $ of $M_{i}$ airways intersect at a point $\mathbf{o}_{IT,i},$
$j=1,\cdots,M_{i},$ namely%
\[
\mathbf{o}_{IT,i}=\mathbf{p}_{\text{e},i_{1}}=\cdots=\mathbf{p}_{\text{e}%
,i_{M_{i}}}%
\]
where $i_{j}$ is the index of airways. The intersection is denoted by\emph{
}$I_{i_{1},i_{2},\cdots,i_{M_{i}}},$ which is a set of a cylinder with center
$\mathbf{o}_{IT,i},$ radius $r_{IT,i}$ and height $h_{IT,i}\geq h_{AW},$
$M_{i}\in%
%TCIMACRO{\U{2124} }%
%BeginExpansion
\mathbb{Z}
%EndExpansion
_{+}.$ We hope that the minimum distance between any pair of airways outside
of an intersection should be greater than $r_{\text{a}}$. Moreover, the
turning radius for all UAVs is $r_{\text{t}}>0.$ For the $i_{j}$th and the
$i_{k}$th airways, the requirement can be formulated as%
\begin{equation}
d\left(  W_{\text{out},i_{j}},W_{\text{out},i_{k}}\right)  >r_{\text{a}},k\neq
j \label{mathinterairway}%
\end{equation}
where
\begin{align*}
W_{\text{out},i_{j}}  &  =W_{\left[  \mathbf{p}_{\text{s},i_{j}}%
,\mathbf{p}_{\text{e},i_{j}}\right]  }\cap \bar{I}_{i_{1},i_{2},\cdots
,i_{M_{i}}}\\
W_{\text{out},i_{k}}  &  =W_{\left[  \mathbf{p}_{\text{s},i_{k}}%
,\mathbf{p}_{\text{e},i_{k}}\right]  }\cap \bar{I}_{i_{1},i_{2},\cdots
,i_{M_{i}}}%
\end{align*}
and $\bar{I}_{i_{1},i_{2},\cdots,i_{M_{i}}}=%
%TCIMACRO{\U{211d} }%
%BeginExpansion
\mathbb{R}
%EndExpansion
^{3}-I_{i_{1},i_{2},\cdots,i_{M_{i}}}$ is the complementary set of
$I_{i_{1},i_{2},\cdots,i_{M_{i}}}.$ For\emph{ }hub intersections and
connections, we have the result shown in \textit{Proposition 3}.

\textbf{Proposition 3}. For the $i$th intersection $I_{i_{1},i_{2}%
,\cdots,i_{M_{i}}}$, line segments $\left[  \mathbf{p}_{\text{s},i_{j}%
},\mathbf{p}_{\text{e},i_{j}}\right]  $ are parallel to the level ground and
at the same altitude, $j=1,\cdots,M_{i}$. If the intersection radius satisfies%
\begin{equation}
r_{IT,i}>\underset{k\neq j,k,j=1,\cdots,M_{i}}{\max}\left(  \frac{r_{\text{a}%
}+\sqrt{h_{AW}^{2}+\left(  2r_{AW}+r_{IS}\right)  ^{2}}}{2\cos \theta_{i}%
\sin \frac{\theta_{i_{j_{i}}i_{j_{k}}}}{2}},r_{\text{t}}\right)
\label{conditionP3}%
\end{equation}
then (\ref{mathinterairway}) holds and the turning radius is satisfied, where%
\begin{align}
\theta_{i_{j_{i}}i_{j_{k}}}  &  =\arccos \frac{\left(  \mathbf{p}%
_{\text{e},i_{j_{i}}}-\mathbf{p}_{\text{s},i_{j_{i}}}\right)  ^{\text{T}%
}\left(  \mathbf{p}_{\text{e},i_{j_{k}}}-\mathbf{p}_{\text{s},i_{j_{k}}%
}\right)  }{\left \Vert \mathbf{p}_{\text{e},i_{j_{i}}}-\mathbf{p}%
_{\text{s},i_{j_{i}}}\right \Vert \left \Vert \mathbf{p}_{\text{e},i_{j_{k}}%
}-\mathbf{p}_{\text{s},i_{j_{k}}}\right \Vert }\label{angletwow}\\
\theta_{i}  &  =\arcsin \frac{r_{AW}+r_{IS}\left/  2\right.  }{r_{IT,i}},k\neq
j,k,j=1,\cdots,N. \label{angletwowa}%
\end{align}

\textit{Proof}. Without loss of generality, we take $W_{\left[  \mathbf{p}%
_{\text{s},i_{1}},\mathbf{p}_{\text{e},i_{1}}\right]  }$ and $W_{\left[
\mathbf{p}_{\text{s},i_{2}},\mathbf{p}_{\text{e},i_{2}}\right]  }$ as an
example. As shown in Fig. \ref{Intersection}, $\mathbf{p}_{\text{e},i_{1}%
,l}^{\prime},$ $\mathbf{p}_{\text{e},i_{1},r}^{\prime},$ $\mathbf{p}%
_{\text{e},i_{1}}^{\prime}$ are all on the middle level plane containing
$\left[  \mathbf{p}_{\text{s},i_{1}},\mathbf{p}_{\text{e},i_{1}}\right]  .$
First, the $i_{1}$th and $i_{2}$th airways can be divided into two parts as%
\begin{align*}
W_{\left[  \mathbf{p}_{\text{s},i_{1}},\mathbf{p}_{\text{e},i_{1}}\right]  }
&  =W_{\left[  \mathbf{p}_{\text{s},i_{1}},\mathbf{p}_{\text{e},i_{1}}%
^{\prime}\right]  }\cup W_{\left[  \mathbf{p}_{\text{e},i_{1}}^{\prime
},\mathbf{p}_{\text{e},i_{1}}\right]  }\\
W_{\left[  \mathbf{p}_{\text{s},i_{2}},\mathbf{p}_{\text{e},i_{2}}\right]  }
&  =W_{\left[  \mathbf{p}_{\text{s},i_{2}},\mathbf{p}_{\text{e},i_{2}}%
^{\prime}\right]  }\cup W_{\left[  \mathbf{p}_{\text{e},i_{2}}^{\prime
},\mathbf{p}_{\text{e},i_{2}}\right]  }.
\end{align*}
Obviously, $W_{\left[  \mathbf{p}_{\text{e},i_{1}}^{\prime},\mathbf{p}%
_{\text{e},i_{1}}\right]  }\subseteq I_{i_{1},i_{2},\cdots,i_{M_{i}}}$
and$\ W_{\left[  \mathbf{p}_{\text{e},i_{2}}^{\prime},\mathbf{p}%
_{\text{e},i_{2}}\right]  }\subseteq I_{i_{1},i_{2},\cdots,i_{M_{i}}}.$ Since%
\begin{equation}
W_{\text{out},i_{1}} \subset W_{\left[  \mathbf{p}_{\text{s},i_{1}}^{\prime
},\mathbf{p}_{\text{e},i_{1}}\right]  }, W_{\text{out},i_{2}} \subset
W_{\left[  \mathbf{p}_{\text{s},i_{2}}^{\prime},\mathbf{p}_{\text{e},i_{2}%
}\right]  }%
\end{equation}
the inequality%
\begin{equation}
d\left(  W_{\left[  \mathbf{p}_{\text{s},i_{1}}^{\prime},\mathbf{p}%
_{\text{e},i_{1}}\right]  },W_{\left[  \mathbf{p}_{\text{s},i_{2}}^{\prime
},\mathbf{p}_{\text{e},i_{2}}\right]  }\right)  >r_{\text{a}} \label{intermid}%
\end{equation}
implies%
\begin{equation}
d\left(  W_{\text{out},i_{1}},W_{\text{out},i_{2}}\right)  >r_{\text{a}}.
\label{requirement}%
\end{equation}
Based on this, we focus on the proof of the condition (\ref{intermid}).
According to \textit{Proposition 2}, we can conclude this proof. $\square$

\textbf{Remark 1}. From condition (\ref{conditionP3}), $\theta_{i_{j_{i}%
}i_{j_{k}}},r_{\text{a}},h_{AW},r_{AW},r_{IS}$ are constants, and $\cos
\theta_{i}\rightarrow1$ as $r_{IT,i}$ is increasing according to
(\ref{angletwowa}). Therefore, $r_{IT,i}$ can always be found for
(\ref{conditionP3}) if it is large enough. To avoid an $r_{IT,i}$ which is too
large, $\theta_{i_{j_{i}}i_{j_{k}}}$ should not be too small.

\subsubsection{Highway Mode for a Connection}

\begin{figure}[b]
\centering
\includegraphics[scale= 1]{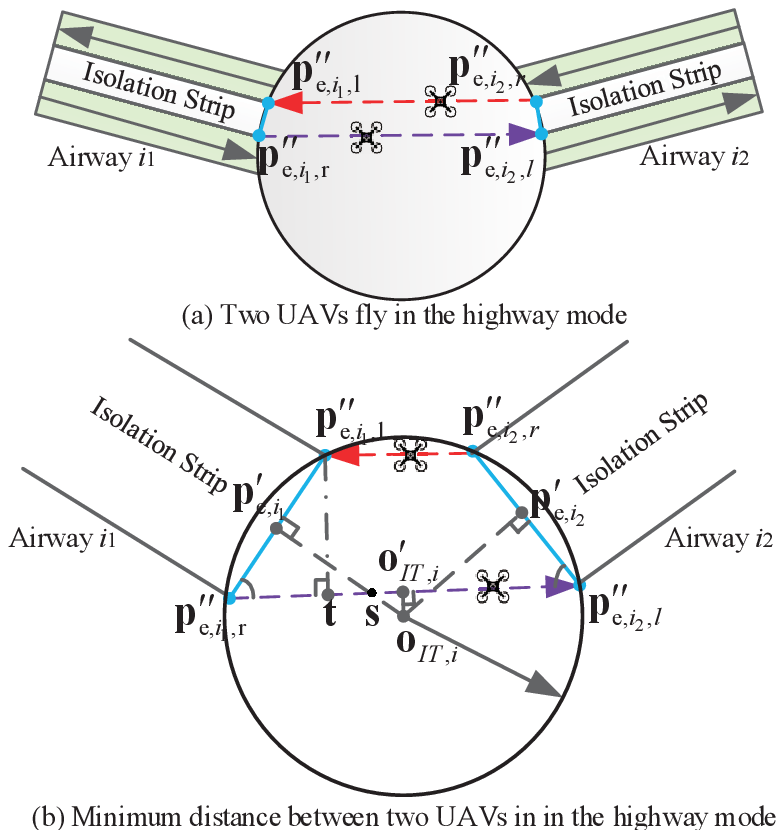} \caption{A connection}%
\label{Connection}%
\end{figure}

It can be considered that two carriageways for traffic traveling in opposite
directions are established in connection. So, basic principles are satisfied
in the connection. Similarly, as shown in Fig. \ref{intersectionnew}(a), all
UAVs fly towards their next airways in the highway mode. For example, as shown
in Fig. \ref{Connection}, we hope that the minimum distance between any pair
of UAVs inside of an intersection should be separated at $r_{\text{a}}$. Let
$\mathbf{p}_{\text{e},i_{j},l}^{\prime \prime}$ and $\mathbf{p}_{\text{e}%
,i_{j},r}^{\prime \prime}$ be the intersection points of the intersection
$I_{i_{1}i_{2}}$ and the isolation strip of the $i_{j}$th airway on the middle
level plane containing $\left[  \mathbf{p}_{\text{s},i_{j}},\mathbf{p}%
_{\text{e},i_{j}}\right]  $, $j=1,2$. Obviously, the distance between line
segment $[ \mathbf{p}_{\text{e},i_{1},l}^{\prime \prime},\mathbf{p}%
_{\text{e},i_{2},r}^{\prime \prime}] $ and line segment $[ \mathbf{p}%
_{\text{e},i_{1},r}^{\prime \prime},\mathbf{p}_{\text{e},i_{2},l}^{\prime
\prime}] $ is the minimum distance for any pair of UAVs from two different
carriageways inside an intersection. If such a distance is greater than
$r_{a},$ then the distance between any pair of UAVs from two different
carriageways inside an intersection is greater than $r_{a}.$ The result is
given in \textit{Proposition 4}.

\textbf{Proposition 4}. For the azimuth connection $I_{i_{1}i_{2}}\ $shown in
Fig. \ref{Connection}, if
\begin{equation}
r_{IS}>\frac{r_{\text{a}}}{\sin \frac{\theta_{i_{1}i_{2}}}{2}} \label{condp4}%
\end{equation}
then%
\begin{equation}
d\left(  \left[  \mathbf{p}_{\text{e},i_{1},l}^{\prime \prime},\mathbf{p}%
_{\text{e},i_{2},r}^{\prime \prime}\right]  ,\left[  \mathbf{p}_{\text{e}%
,i_{1},r}^{\prime \prime},\mathbf{p}_{\text{e},i_{2},l}^{\prime \prime}\right]
\right)  >r_{\text{a}}. \label{resultdp4}%
\end{equation}

\emph{Proof}. First, it is obvious that $[ \mathbf{p}_{\text{e},i_{1}%
,l}^{\prime \prime},\mathbf{p}_{\text{e},i_{2},r}^{\prime \prime}] \parallel[
\mathbf{p}_{\text{e},i_{1},r}^{\prime \prime},\mathbf{p}_{\text{e},i_{2}%
,l}^{\prime \prime}] .$ So,
\begin{align}
&  d\left(  \left[  \mathbf{p}_{\text{e},i_{1},l}^{\prime \prime}%
,\mathbf{p}_{\text{e},i_{2},r}^{\prime \prime}\right]  ,\left[  \mathbf{p}%
_{\text{e},i_{1},r}^{\prime \prime},\mathbf{p}_{\text{e},i_{2},l}^{\prime
\prime}\right]  \right) \nonumber \\
&  =d\left(  \mathbf{p}_{\text{e},i_{1},l}^{\prime \prime},\left[
\mathbf{p}_{\text{e},i_{1},r}^{\prime \prime},\mathbf{p}_{\text{e},i_{2}%
,l}^{\prime \prime}\right]  \right) \nonumber \\
&  =r_{IS}\sin \angle \mathbf{p}_{\text{e},i_{1},l}^{\prime \prime}%
\mathbf{p}_{\text{e},i_{1},r}^{\prime \prime}\mathbf{p}_{\text{e},i_{2}%
,l}^{\prime \prime}. \label{equ1p4}%
\end{align}
Since the triangle $\triangle \mathbf{p}_{\text{e},i_{1},r}^{\prime \prime
}\mathbf{sp}_{\text{e},i_{1}}^{\prime}$ is similar to the triangle
$\triangle \mathbf{o}_{IT,i}\mathbf{so}_{IT,i}^{\prime},$ we have%
\begin{equation}
\angle \mathbf{p}_{\text{e},i_{1},l}^{\prime \prime}\mathbf{p}_{\text{e}%
,i_{1},r}^{\prime \prime}\mathbf{p}_{\text{e},i_{2},l}^{\prime \prime}%
=\frac{\theta_{i_{1}i_{2}}}{2} \label{anglep4}%
\end{equation}
where $\theta_{i_{1}i_{2}}$ is shown in Fig. \ref{Intersection}(b). According
to (\ref{equ1p4}) and (\ref{anglep4}), if (\ref{condp4}) is satisfied, then
(\ref{resultdp4}) holds. $\square$

\subsubsection{Rotary Island Mode for a Hub}

A rotary island as shown in Fig. \ref{intersectionnew}(b) can be considered as
a bended carriageway. In the rotary island mode, all UAVs entering the
intersection will perform a clockwise or anticlockwise rotation until they fly
at the entrance to the next airway. Other operations are similar to those in
the highway mode. Hence, basic principles can be satisfied in a hub. One
intersection can take several UAVs in it at the same time. Applying the
highway mode to a hub directly will cause the jam problem because of the
difference among the flight directions of UAVs in hub. In the rotary island
mode, the velocity of a UAV aligns with those around it, which is similar to
the situation of birds flying in the sky. It allows more UAVs in the hub
simultaneously, in addition to make it easier to avoid jam.

\section{EXPERIMENTS AND RESULTS}

We design a set of experiments to demonstrate the effectiveness of the
designed sky highway above. The video is available at https://youtu.be/WDhNroCMD04.

\subsection{Simulation Experiment}

Controller design is based on Matlab R2017b and Simulink 9.3, which is used
for generating speed command of every UAV based on the artificial potential
field method \cite{Quan(2011)}. The model simulator configures parameters of
UAVs, and generate sensor data to the controller via UDP communication
network. The 3D viewer is based on Unreal Engine 4, which is used for
displaying the real-time flight performance with high-quality
screens.\begin{figure}[t]
\centering
\includegraphics[scale=0.61]{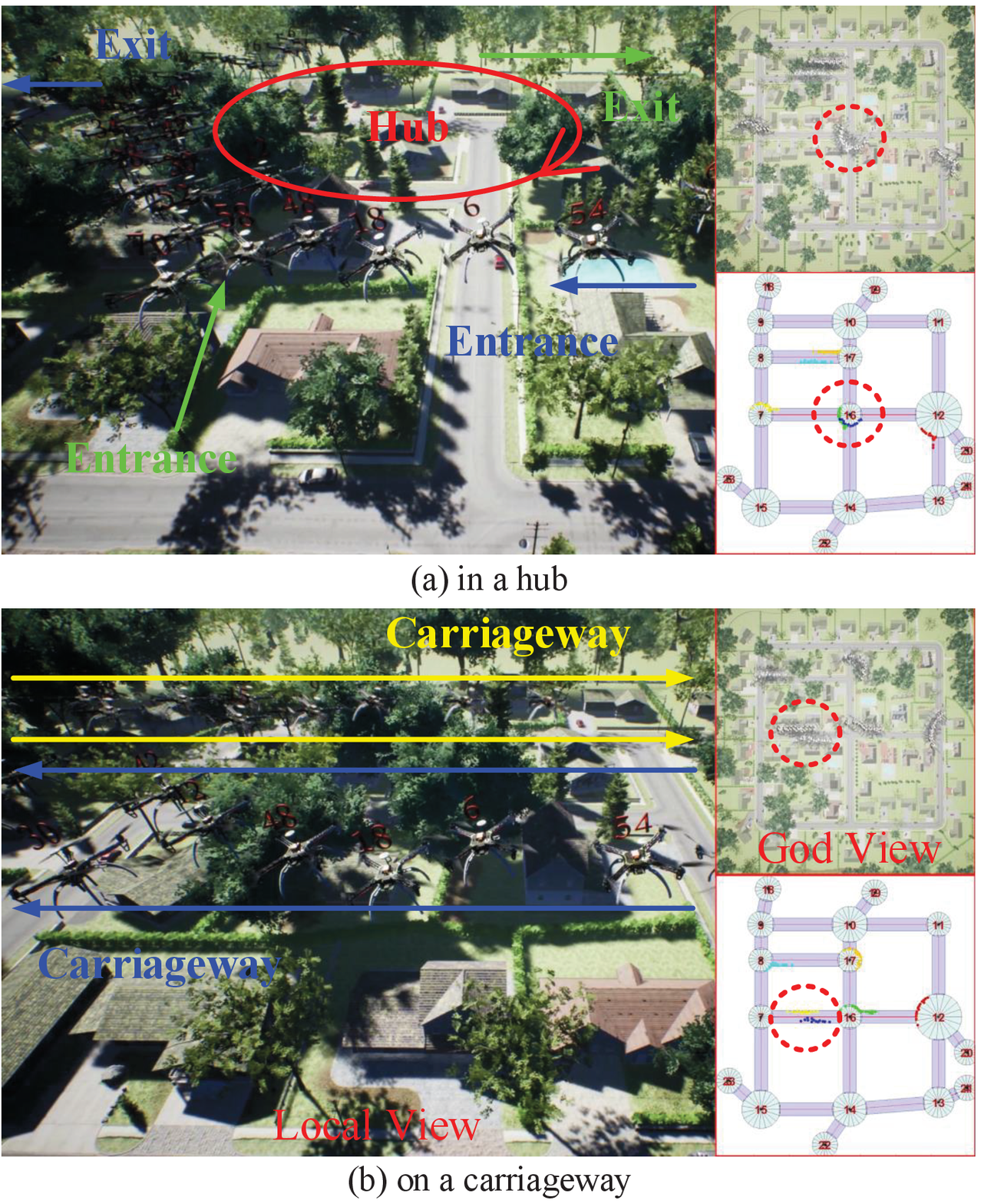} \caption{Sky highway
simulation}%
\label{SimulationResults}%
\end{figure}

Build a traffic network shown in Fig. \ref{SimulationResults}, assuming that
$r_{\text{a}}=3$ m, $r_{AW}=h_{AW}=9$ m. There are eighty UAVs with different
routes and colors. Each UAV is required to depart from its airport and then
pass the nodes in its route according to its specified sequence. During the
whole process, UAVs are only permitted to fly within airways.

\begin{figure}[t]
\centering
\includegraphics[scale= 1]{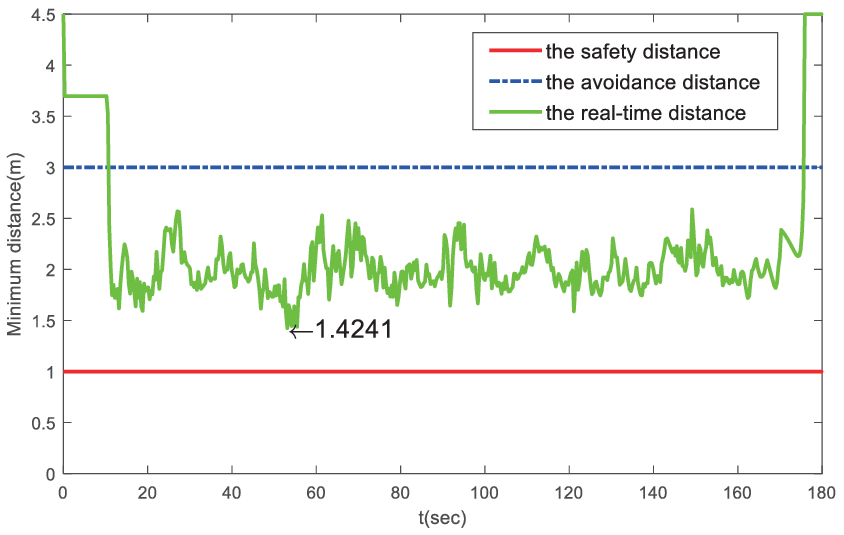} \caption{The minimum
distance among all UAVs during the whole simulation}%
\label{SimulationDistance}%
\end{figure}

As shown in Fig. \ref{SimulationResults}, the airways and intersections are
designed within the geometrical structure of the sky highway generation
module. The width $r_{IS}$ of each isolation strip satisfies conditions
(\ref{p1con}) and (\ref{p2con}), and airways connected with azimuth
connections satisfy the condition (\ref{condp4}). In addition, each
intersection radius satisfies the condition (\ref{conditionP3}). As shown in
Fig. \ref{SimulationResults}(a), we track UAVs labeled blue color. UAVs in
blue and green all enter Node $16$ and leave through different carriageways.
They perform a clockwise rotation in rotary island mode so that the UAV
traffic flow can be increased greatly. As shown in Fig.
\ref{SimulationResults}(b), UAVs in blue enter the carriageway connecting Node
$16$ to Node $7$ in the highway mode. Unlike in line, some UAVs fly neck by
neck, so the airspace is utilized sufficiently in the carriageway. Meanwhile,
UAVs in blue do not affect the UAVs in yellow on the other carriageway in the
same airway because of the isolation strip. Finally, all UAVs complete their
routes, and they always keep a safe distance from others as shown in Fig.
\ref{SimulationDistance}.

\subsection{Flight Experiment}

\begin{figure}[t]
\centering
\includegraphics[scale= 0.45]{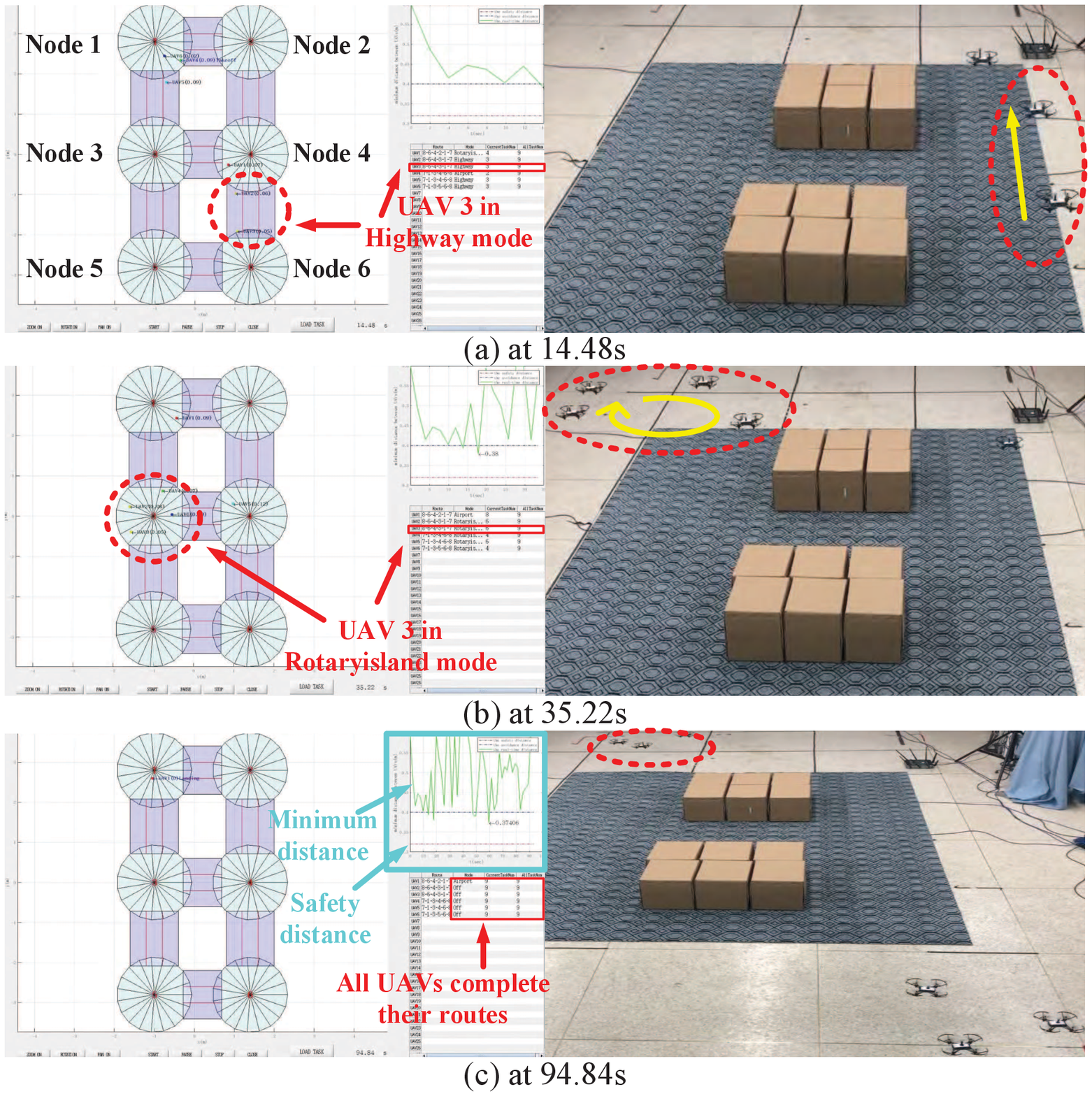} \caption{The position, velocity and
mode of each UAV during the whole flight experiment}%
\label{Simulation}%
\end{figure}

An experiment is carried out in a laboratory room with an OptiTrack motion
capture system installed, which provides the ground truth of the positions and
orientations of UAVs. The laptop mainly includes four modules: the geometrical
structure of the sky highway generation, flight mode switching, plotting and
UAV control. Build a traffic network shown in Fig. \ref{Simulation}, assuming
that $r_{\text{a}}=0.4$ m, $r_{AW}=h_{AW}=0.6$ m, $\mathbf{o}_{IT,1}=\left[
{-1\; \;2.8\; \;1}\right]  ^{\text{T}}$ m, $\mathbf{o}_{IT,2}=\left[  {1.4\;
\;2.8\; \;1}\right]  ^{\text{T}}$ m, $\mathbf{o}_{IT,3}=\left[  {-1\; \;0\;
\;1}\right]  ^{\text{T}}$ m, $\mathbf{o}_{IT,4}=\left[  {1.4\; \;0\;
\;1}\right]  ^{\text{T}}$ m, $\mathbf{o}_{IT,5}=\left[  {-1\; \;-2.8\;
\;1}\right]  ^{\text{T}}$ m, $\mathbf{o}_{IT,6}=\left[  {1.4\; \;-2.8\;
\;1}\right]  ^{\text{T}}$ m. Add Node $7$ and Node $8$ connected with Node $1$
and Node $6$ as airports with $\mathbf{o}_{IT,7}=\left[  {-1\; \;2.8\;
\;0}\right]  ^{\text{T}}$ m and $\mathbf{o}_{IT,8}=\left[  {1.4\; \;-2.8\;
\;0}\right]  ^{\text{T}}$ m, respectively. There are six UAVs with different
routes and colors.

We track UAV $3$ by the dotted circle as shown in Fig. \ref{Simulation}(a).
After take-off, it enters carriageway connecting Node $6$ to Node $4$ in the
highway mode at 14.48 second. As shown in Fig. \ref{Simulation}(b), Node $3$
is a hub, and UAVs $2$, $4$ and $6$ also enter this intersection at 35.22
second. They all perform a clockwise rotation in rotary island mode. Finally,
all UAVs complete their routes at 94.84 second, and they always keep a safe
distance from others as shown in Fig. \ref{Simulation}(c).

\section{CONCLUSIONS}

The geometrical structures of airways and intersections are designed to
maintain a certain distance between two UAVs on two carriageways. For the
situation that two carriageway in one airway, condition (\ref{p1con}) has to
be satisfied while for the situation that two carriageways in two airways,
condition (\ref{p2con}) has to be satisfied. The highway mode is applied to
airways for dense traffic. As more than one airways will be connected with one
intersection, the radius of the intersection should be large enough so that
two airways outside of the intersection can keep a specified distance. For
a\emph{ }hub or a connection, condition (\ref{conditionP3}) has to be
satisfied. For dense traffic, the proposed highway mode is applied to
connection intersections while the proposed rotary island mode is for hub
intersections. Finally, the effectiveness of the proposed sky highway is shown
by the given demonstration.


\begin{thebibliography}{99}                                                                                               %


\bibitem {Balakrishnan(2018)}K. Balakrishnan, J. Polastre, J. Mooberry, R.
Golding and P. Sachs, Premiering a future blueprint for our sky, available at
https://www.utmblueprint.com, February 28, 2019.

\bibitem {NASA}NASA, UAS Traffic Management (UTM), available at
https://utm.arc.nasa.gov, February 28, 2019.

\bibitem {SESAR}SESAR, European Drones Outlook 2016, available at
https://www.sesarju.eu/sites/default/files/documents/reports/European$\_$
Drones$\_$Outlook$\_$Study$\_$2016.pdf, February 28, 2019.

\bibitem {FreeFlight(1995)}RTCA, Final report of the RTCA Task Force 3, free
flight implementation, RTCA, Inc., Washington, DC, Oct 26, 1995

\bibitem {Jana(1997)}J. Kosecka, C. Tomlin, G. Pappas, et al, Generation of
conflict resolution maneuvers for air traffic management, \emph{International
Conference on Intelligent Robots and Systems (IROS)}, 1997, pp 1598-1603.

\bibitem {ICAO(2012)}ICAO, The use of displayed ADS-B data for a collision
avoidance capability in unmanned aircraft system, UASSG/9-SN, no. 5, 2012.

\bibitem {Hayat(2016)}S. Hayat, E. Yanmaz and R. Muzaffar, Survey on unmanned
aerial vehicle networks for civil applications: A communications viewpoint,
\emph{IEEE Communications Surveys \& Tutorials}, vol. 18, no. 4, 2016, pp 2624-2661.

\bibitem {GSMA(2018)}GSMA, Mobile-enabled unmanned aircraft, available at
https://www.gsma.com/iot/wp-content/uploads/2018/02/Mobile-Enabled-Unmanned-Aircraft-web.pdf,
February 28, 2019.

\bibitem {IoD(2016)}M. Gharibi, R. Boutaba and S. L. Waslander, Internet of
Drones, \emph{IEEE Access}, vol. 4, 2016, pp 1148-1162.

\bibitem {Devasia(2016)}S. Devasia and A. Lee, A scalable low-cost-UAV traffic
network (uNet), \emph{Journal of Air Transportation}, vol. 24, no. 3, 2016, pp 74-83.

\bibitem {Mcfadyen(2017)}A. Mcfadyen and T. Bruggemann, Unmanned air traffic
network design concepts, \emph{2017 IEEE 20th International Conference on
Intelligent Transportation Systems (ITSC)}, 2017, pp 1-7.

\bibitem {Viragh(2016)}C. Vir\'{a}gh, M. Nagy, C. Gershenson and G.
V\'{a}s\'{a}rhelyi, Self-organized UAV traffic in realistic environments,
\emph{2016 IEEE/RSJ international conference on intelligent robots and systems
(IROS)}, 2016, pp 1645-1652.

\bibitem {Boldizsar(2018)}B. Bal\'{a}zs and G. V\'{a}s\'{a}rhelyi, Coordinated
dense aerial traffic with self-driving drones, \emph{2018 IEEE International
Conference on Robotics and Automation (ICRA)}, 2018, pp 6365-6372.

\bibitem {Chung(2018)}S. J. Chung, A. A. Paranjape, P. Dames, S. Shen and V.
Kumar, A survey on aerial swarm robotics, \emph{IEEE Transactions on
Robotics}, vol. 34, no. 4, 2018, pp. 837-855.

\bibitem {Dutta(2010)}K. Dutta, How birds fly together: The dynamics of
flocking, \emph{Resonance}, vol. 15, no. 12, 2010, pp. 1097-1110.

\bibitem {Quan(2011)}Quan Q, \emph{Introduction to Multicopter Design and
Control}. Singapore: Springer Singapore, 2017.
\end{thebibliography}
\end{document}